# Syntax-based default reasoning as probabilistic model-based diagnosis

Jérôme Lang
IRIT
Université Paul Sabatier
31062 Toulouse Cedex
France
email: lang@irit.fr

## Abstract

We view the syntax-based approaches to default reasoning as a model-based diagnosis problem, where each source giving a piece of information is considered as a component. It is formalized in the ATMS framework (each source corresponds to an assumption). We assume then that all sources are independent and "fail" with a very small probability. This leads to a probability assignment on the set of candidates, or equivalently on the set of consistent environments. This probability assignment induces a Dempster-Shafer belief function which measures the probability that a proposition can be deduced from the evidence. This belief function can be used in several different ways to define a nonmonotonic consequence relation. We study ans compare these consequence relations. The case of prioritized knowledge bases is briefly considered.

## 1 Introduction

Syntax-based approaches to inconsistency handling, default reasoning and belief revision have been proposed and studied in various forms (e.g. [14], [16], [31], [15], [4], [6], [24], and especially [25] and [1]). They assume that the input (the *knowledge base* - KB for short) consists of a set of logical sentences, possibly equipped with a priority ordering; when this knowledge base is inconsistent, these approaches select among the consistent sub-bases of the KB some *preferred* sub-bases (the selection criterion can be for instance maximality w.r.t. set inclusion[1] or cardinality). A consequence relation is then generally defined by taking the intersection of the logical closures of these preferred sub-bases. Each formula of the KB is considered as a distinct piece of information, which can be kept in the knowledge base or rejected from it independently from the others; therefore, it may happen that two semantically equivalent knowledge bases may be revised differently and thus lead to different conclusions - this is why they are called *syntax-based*. Consider for instance $K_1 = \{p, \neg p, q\}$ and $K_2 = \{p \wedge q, \neg p\}$; $q$ holds in all maximal consistent sub-bases of $K_1$ but this is not the case for $K_2$. When cardinality is used to select preferred subbases, even the number of occurrences of identical (or logically equivalent) formulas in $K$ matters: for instance, $\{p, \neg p\}$ has two consistent sub-bases of maximum cardinality ($\{p\}$ and $\{\neg p\}$) whereas $\{p, \neg p, \neg p\}$ has only one ($\{\neg p, \neg p\}$).

Now, in model-based diagnosis, the *consistency-based approaches* (see [37], [13], [35]) proceed in a very similar manner, since they look for *preferred candidates*, i.e. minimal (w.r.t. a given selection criterion) sets of faulty components, such that the description of how the non-faulty components work is consistent with the observations[2]. The link between default reasoning and model-based diagnosis has already been well studied (e.g. [33], [37], [26], [20]): indeed, the principles behind consistency-based diagnosis and syntax-based approaches are basically the same: there is a correspondance between a source providing us with a piece of information and a component of a diagnosis problem; a faulty component corresponds to an erratic source which gives a piece of information which is not relevant (by analogy, we will say that the source is *faulty*). When the component is working correctly, the formula describing its normal behaviour must be satisfied, and analogously, when the source is not faulty, the formula associated to it must be true in the real world. Then, a candidate in a diagnosis problem (i.e. a set of components consistent with the observations) corresponds to a candidate in a syntax-based default reasoning problem (i.e. a set of formulas whose deletion restores the consistency of the knowledge base).

In the well-known diagnosis system GDE, De Kleer and Williams [11] propose a probabilistic criterion to rank candidates: each component has an initial probability of fault, and it is assumed that components fail independently; then, the a posteriori probability that a given candidate is the *real* candidate is computed via Bayes' rule, conditioning by the observations. This principle of ranking candidates w.r.t. their probability assumes the initial probabilities of fault are available. When it is not the case, De Kleer [12] proposes to assume that all components have a very small probability of fault. What we propose to do here is to use a similar assumption for syntax-based default reasoning, which induces probabilities of the consistent sub-bases of the KB (which comes down to compute the probabilities of the candidates - a candidate specifies which pieces of information have to be rejected and thus which ones remain in the KB). We will check that, as expected, the consistent sub-bases of maximal cardinality are the most probable ones. This probability distribution induces then a Dempster-Shafer belief function, which evaluates the probability that a formula can be proved from the available evidence (which consists only in the KB and the assumptions of independence and small probabilities of fault). The most original contribution of this paper is to propose (and to compare) many different ways to define a syntax-based consequence relation from this induced belief function. An interesting point is that we will then recover some already known syntax-based consequence relations (but with a new justification) and obtain a few new ones. Lastly, we propose briefly a generalization to the case of prioritized knowledge bases.

---

[1] In this case the preferred sub-bases coincide with the extensions of default logic [36] restricted to normal defaults without prerequisites.

[2] The principle of minimizing the set of faulty components w.r.t. a given criterion is generally called *principle of parsimony* (see e.g. [29]).

## 2   Inconsistent knowledge bases as systems to diagnose

From now on, $\mathcal{L}$ denotes a propositional language generated by a finite number of propositional variables. Formulas will be denoted by greek letters $\varphi$, $\psi$, etc. $\top$ denotes tautology, $\models$ classical entailment and $Cn$ logical closure.

A *knowledge base* (KB) intuitively consists of a set $\mathcal{F}$ of hard facts which cannot be rejected, and a *multiset* $\Delta$ of default formulas which can be rejected if necessary[3]. To distinguish each default from the others, we create a set of assumptions $\mathcal{A} = \{A_1, ..., A_n\}$ (with as many assumptions as defaults) and label each default with a distinct assumption. We define a knowledge base as in [31] and we then recall well-known definitions of the ATMS and model-based diagnosis literatures [10], [37], [11], [13].

**Definition 1** *A* **knowledge base** $K$ *is defined as a couple* $K = (\mathcal{F}, \Delta)$ *where*

- $\mathcal{F}$ *is a finite set of formulas (hard facts)*

- $\Delta = \{\varphi_1, ..., \varphi_n\}$ *a finite multiset of formulas (defaults).*

*The* **assumption set** $\mathcal{A}(K)$ *associated to* $K$ *(denoted by* $\mathcal{A}$ *when no confusion is possible) is defined by* $\mathcal{A} = \{A_1, ..., A_n\}$ *where each assumption is associated to a default[4] by the mapping* $\delta$: $\forall i = 1...n$, $\delta(A_i) = \varphi_i$.

**Definition 2** *A subset of* $\mathcal{A}$ *is called an* **environment**. *The* **context** *of an environment* $E$ *is defined by* $Context(E) = Cn(\mathcal{F} \cup \{\varphi_i | A_i \in E\})$[5]. *An environment* $E$ *is* **consistent** *iff* $Context(E)$ *is consistent. It is* **irredundant** *iff no proper superset of* $E$ *is consistent*[6]. *It is consistent with maximal cardinality (or, for short,* **maxcard consistent***) iff for any consistent* $E'$ *we have* $|E| \geq |E'|$.
*A* **nogood** *is an inconsistent environment. A* **candidate** $C$ *is the complementary of a consistent environment. It is* **minimal** *iff no proper subset of* $C$ *is a candidate; it is a candidate of minimal cardinality (or* **mincard** *for short) iff for any candidate* $C'$ *we have* $|C| \leq |C'|$[7].

Pursuing the analogy with model-based diagnosis, the source of information corresponding to the assumption $A_i$ can be viewed as a component; $\varphi_i$ is then the logical description of how the component works. If $A_i$ is true then the source is "non-faulty" and the associated formula $\varphi_i$ is satisfied in the real world; if $A_i$ is false then the source is "faulty" and then we don't know whether the associated formula is satisfied or not in the real world (in terms of diagnosis, it corresponds to the assumption that we don't know how behaves a faulty component).

As in [13] a nogood $\{A_{i_1}, ..., A_{i_p}\}$ will also be written logically by $\neg A_{i_1} \vee ... \vee \neg A_{i_p}$[8]; a candidate $\{A_{j_1}, ..., A_{j_q}\}$ will also be written logically by $\neg A_{j_1} \wedge ... \wedge \neg A_{j_q}$. The nogood base, denoted by $\neg N$, is the conjunction of all irredundant nogoods; it is well-known to be equivalent to the conjunction of all minimal nogoods, and as well to the disjunction of all [irredundant] candidates [13]. A detailed example is given in Section 3 and continued in Section 4.

---

[3] We recall that in a multiset several occurrences of the same element are distinguished: this obviously has to be the case for syntax-based approaches where several occurrences of the same default constitute several distinct pieces of information.

[4] Instead of this we could have equivalently generated the set of ATMS justifications $A_i \rightarrow \varphi_i$

[5] Note that $Context(\mathcal{A}) = K$.

[6] This is called an *interpretation* in [10]

[7] Obviously, a minimal (resp. mincard) candidate is the complementary of an irredundant (resp. maxcard) consistent environment.

[8] Note that $\neg A_{i_j}$ corresponds to De Kleer et al.'s [13] notation $AB(c_{i_j})$ meaning that the component $c_{i_j}$ is faulty.

# 3 From syntactical knowledge bases to belief functions

## 3.1 Computing the probability of environments

As in [12] we make the two following basic assumptions:

- **(I)** each assumption is *independent* from the others. This means that each default piece of information is kept or rejected independently from the others - which is in accordance with the spirit of syntax-based approaches to default reasoning.

- **(S)** all assumptions are assigned the same initial probability (the sources have the same prior probability of fault), and this probability of fault is very small: $\forall i, Prob(\neg A_i) = \varepsilon$, with $\varepsilon \ll 1$.

This leads to define a probability assignment on the environment set $2^\mathcal{A}$. Thus, the prior probability of an environment $E$ of cardinality $k$ is $Pr(E) = \varepsilon^{n-k}(1-\varepsilon)^k$ (which is approximated by $\varepsilon^{n-k}$ when $\varepsilon \to 0$). $Pr(E)$ is the prior probability that $E$ is the *real* environment, i.e. the environment corresponding to the real world. Now, this real environment must be consistent; to ensure that inconsistent environments are given a zero probability, the prior probability is conditioned on the consistent environments (see e.g. [22]), i.e.

$$Pr(E|\neg N) = \frac{Pr(E \wedge \neg N)}{Pr(\neg N)}$$

**Proposition 1** *Assume that there are exactly $p$ maxcard consistent environments; let $k$ be their cardinality. Let $E$ be any consistent environment. Then*[9]

- *if $|E| = k$ then $Pr(E|\neg N) = \frac{1}{p} + O(\varepsilon)$*

- *if $|E| < k$ then $Pr(E|\neg N) = O(\varepsilon^{k-|E|})$*

Proof: let us prove first prove that $Pr(\neg N) = p\varepsilon^{n-k} + O(\varepsilon^{n-k+1})$. Let $C_1, ..., C_p$ be the mincard candidates; they are the complementary of the maxcard consistent environments, so their cardinality is $n - k$. Let $C_{p+1}, ..., C_q$ be the other irredundant candidates. $Pr(\neg N) = Pr(C_1 \vee ... \vee C_q) = Pr(C_1) + ... + Pr(C_p) + Pr(C_{p+1}) + ... + Pr(C_q) - \sum_{i \neq j} Pr(C_i \wedge C_j) + \sum_{i \neq j, j \neq l, i \neq l} Pr(C_i \wedge C_j \wedge C_l) + ...$. Now, $Pr(C_1) = ... = Pr(C_p) = \varepsilon^{n-k}$; $\forall i = p+1...q, Pr(C_i) = \varepsilon^{n-k+1}$; and $\forall i, j$ such that $i \neq j$, $C_i \wedge C_j$ contains at most $n - k + 1$ literals $\neg A_i$'s (if it contained only $n - k$, since $n - k$ is the maximum cardinality of a consistent environment, one of the two candidates $C_i$ and $C_j$ would be contained in the other one, which would contradict the fact they are irredundant); thus, $Pr(C_i \wedge C_j) \leq \varepsilon^{n-k+1}$. A fortiori, for all conjunctions of more than two $C_i$'s: $Pr(C_{i_1} \wedge ... \wedge C_{i_m}) \leq \varepsilon^{n-k+1}$. Thus, $Pr(\neg N) = p\varepsilon^{n-k} + O(\varepsilon^{n-k+1})$. Now, $E \models \neg N$ so $Pr(E|\neg N) = \frac{Pr(E)}{Pr(\neg N)} = \frac{\varepsilon^{n-|E|}(1-\varepsilon^{|E|})}{p\varepsilon^{n-k}+O(\varepsilon^{n-k+1})}$; therefore if $|E| = k$, $Pr(E|\neg N) = \frac{1}{p} + O(\varepsilon)$; and if $|E| < k$, $Pr(E|\neg N) = O(\varepsilon^{k-|E|})$.

Computing the probability of the consistent environments is exactly the same task as computing the probabilities of candidates in consistency-based model-based diagnosis ([11], [12], [35]). Proposition 1 tells that the only consistent environments whose probability does not tend to 0 when $\varepsilon \to 0$ are those of maximal cardinality. This is in accordance with a version of the principle of parsimony consisting in considering only the candidates of minimum cardinality ([12], [29]).

It is also interesting to compute the probability of fault of a single source, namely $Pr(\neg A_i|\neg N)$:

**Proposition 2** *As before, assume that there are exactly $p$ maxcard consistent environments and let $k$ be their cardinality. Let $A_i$ be an assumption.*

- *if $A_i$ is absent of $r \geq 1$ maxcard consistent environments, then*

$$Pr(\neg A_i|\neg N) = \frac{r}{p} + O(\varepsilon)$$

- *if $A_i$ appears in all maxcard consistent environments, and is absent of $r'$ irredundant consistent environments of cardinality $k - 1$, then*

$$Pr(\neg A_i|\neg N) = (1 + \frac{r'}{p})\varepsilon + O(\varepsilon^2)$$

The proof uses the same kind of considerations as the proof of Proposition 1.

*Remark*: if $A_i$ appears in all irredundant consistent environments, then $r' = 0$ and Proposition 1 gives $Pr(\neg A_i|\neg N) = \varepsilon$. Indeed, in this case, $\neg A_i$ never appears in $\neg N$ and therefore $\neg A_i$ and $\neg N$ are independent; thus $Pr(\neg A_i|\neg N) = Pr(\neg A_i) = \varepsilon$.

*Example*: $F = \{a\}$ and $\Delta$ contains the following formulas (with their respective $A_i$'s):

| | |
|---|---|
| $A_1$ | $a \to b \wedge e \wedge f$ |
| $A_2$ | $a \to c \wedge d$ |
| $A_3$ | $\neg b \vee \neg d$ |
| $A_4$ | $e$ |
| $A_5$ | $\neg b \wedge \neg c \wedge \neg e \wedge g$ |
| $A_6$ | $b \wedge \neg c \wedge d \wedge \neg e \wedge \neg g$ |

Here are the irredundant consistent environments, their probability and their context[10]:

---

[9]We recall that the notation $O(\varepsilon^k)$ denotes any function $f$ of $\varepsilon$ such that $\frac{f(\varepsilon)}{\varepsilon^{k-1}} \to_{\varepsilon \to 0} 0$.

[10]We omit the $Cn$ notation in the context culumn, so for instance it should be read $Context(\{A_1, A_2, A_4\}) = Cn(\{a, b, c, d, e, f\})$ etc.

| $E$ | $Pr(E\|\neg N)$ | $Context(E)$ |
|---|---|---|
| $\{A_1, A_2, A_4\}$ | $\frac{1}{3} + O(\varepsilon)$ | $a, b, c, d, e, f$ |
| $\{A_1, A_3, A_4\}$ | $\frac{1}{3} + O(\varepsilon)$ | $a, b, \neg d, e, f$ |
| $\{A_2, A_3, A_4\}$ | $\frac{1}{3} + O(\varepsilon)$ | $a, \neg b, c, d, e$ |
| $\{A_3, A_5\}$ | $O(\varepsilon)$ | $a, \neg b, \neg c, \neg e, g$ |
| $\{A_6\}$ | $O(\varepsilon^2)$ | $a, b, \neg c, d, \neg e, \neg g$ |

Here are the $Pr(\neg A_i|\neg N)$:
$Pr(\neg A_1|\neg N) = Pr(\neg A_2) = Pr(\neg A_3) = \frac{1}{3} + O(\varepsilon)$;
$Pr(\neg A_4|\neg N) = (1 + \frac{1}{3})\varepsilon + O(\varepsilon^2) = \frac{4}{3}\varepsilon + O(\varepsilon^2)$;
$Pr(\neg A_5|\neg N) = Pr(\neg A_6|\neg N) = 1 + O(\varepsilon)$;

The maxcard consistent environments are $\{A_1, A_2, A_4\}$, $\{A_1, A_3, A_4\}$ and $\{A_2, A_3, A_4\}$.

### 3.2 How probabilities of candidates induce a belief function

We have seen that the knowledge base $K$ induces a probability assignment of the environment see $2^A$. This probability assignment of the assumption set induces a Dempster-Shafer belief function (see [22], [34], [27], [9], [38] for a study of this connection between ATMS and belief functions). As studied in detail by Smets [38], this belief function represents a *probability of deductibility*, i.e. the probability that the evidence is sufficient to prove the proposition (see also [22], [27]). This belief function is given by[11]

$$Bel_K(\psi) = \sum_{E \in 2^A, \psi \in Context(E)} Pr(E|\neg N)$$

**Proposition 3** $Bel_K(\psi) = 1$ iff $\mathcal{F} \models \psi$

*Proof:* if $\mathcal{F} \models \psi$ for any environment $E$, $\psi \in Context(E)$ and therefore $Bel_K(\psi) = 1$. Reciprocally, if $Bel_K(\psi) = 1$ then consider the environment $\emptyset$; it has a non-zero probability and its context is only $\mathcal{F}$, therefore $\mathcal{F} \models \psi$.

**Proposition 4** *Let $k_\psi$ be the maximum cardinality of a consistent environment $E$ such that $\psi \in Context(E)$ (if any) and let $u_\psi$ be the number of such environments; as before, let $k$ be the cardinality of a maxcard consistent environment. Then*

- *if $k_\psi = k$ then*

$$Bel_K(\psi) = \frac{u_\psi}{p} + O(\varepsilon)$$

- *if $k_\psi < k$ then*

$$Bel_K(\psi) = O(\varepsilon^{k-k_\psi})$$

---

[11]An equivalent expression of $Bel_K(\psi)$ is (see for instance [22])

$$Bel_K(\psi) = \frac{Pr(label(\psi) \wedge \neg N)}{Pr(\neg N)}$$

where $label(\psi)$ is the logical expression of the set of all irredundant consistent environments in which $\psi$ is provable.

- *if there is no consistent environment $E$ such that $\psi \in Context(E)$ then*

$$Bel_K(\psi) = 0$$

The proof comes immediately from Proposition 1.

*Example* (continued):
$Bel_K(b \vee c) = 1$;
$Bel_K(b) = \frac{2}{3} + O(\varepsilon)$;
$Bel_K(g) = O(\varepsilon)$;
$Bel_K(\neg g) = O(\varepsilon^2)$;
$Bel_K(\neg f) = 0$.

## 4 Inducing consequence relations

We have seen that, given a knowledge base $K$, and assuming small fault probabilities and independence of the sources, we obtain a belief function $Bel_K$ on $\mathcal{L}$ induced by $K$; $Bel_K(\psi)$ is the probability that $\psi$ be deductible from $K$ from the evidence. Now, we can use this generated belief function to define nonmonotonic consequence relations (CR) on $\mathcal{L}$. We are going to investigate several proposals of CRs, many of which will appear to be well-known. We define the CRs in the syntax as Pinkas and Loui [30], namely $K \hspace{1pt}\vdash\hspace{-6pt}\sim\hspace{2pt} \psi$ means that the formula $\psi$ is inferred from the knowledge base $K$[12].

As in [31] we define a **scenario** of $K = (\mathcal{F}, \Delta)$ as a consistent subset $S$ of $\mathcal{F} \cup \Delta$ containing $\mathcal{F}$ (note that $Cn(S)$ is the context of a consistent environment). A scenario is said *irredundant* (resp. *maxcard*) iff it is maximal w.r.t. set inclusion (resp. cardinality).

**Definition 3** $K \hspace{1pt}\vdash\hspace{-6pt}\sim\hspace{2pt}_1 \psi$ iff $Bel_K(\psi) \rightarrow_{\varepsilon \to 0} 1$

**Proposition 5** $K \hspace{1pt}\vdash\hspace{-6pt}\sim\hspace{2pt}_1 \psi$ iff for any maxcard scenario $S$ of $K$, $S \models \psi$.

The proof comes easily from the fact that only maxcard consistent environments have a probability which does not tend to 0. This kind of CR is known as a *strong* CR. More precisely, $\hspace{1pt}\vdash\hspace{-6pt}\sim\hspace{2pt}_1$ has been studied in a more general setting (and with priorities) in [23] and [1][13]. This result gives thus a new interpretation of this well-known inference relation.

---

[12]Note that, in spite of the syntax $K \hspace{1pt}\vdash\hspace{-6pt}\sim\hspace{2pt} \psi$, $\hspace{1pt}\vdash\hspace{-6pt}\sim\hspace{2pt}$ is actually a *unary* CRs; a *binary* CR induced by $K$ would be $\hspace{1pt}\vdash\hspace{-6pt}\sim\hspace{2pt}_K$ where $\varphi \hspace{1pt}\vdash\hspace{-6pt}\sim\hspace{2pt}_K \psi$ means that with respect to the background knowledge represented by $K$, if we assume $\varphi$ then we infer nonmonotonically $\psi$ (and the unary case is recovered when $\varphi = \top$). For the sake of simplicity, in this paper we define only the unary restrictions of the CRs; however this restriction is not essential: indeed, syntax-based CRs generally satisfy $\varphi \hspace{1pt}\vdash\hspace{-6pt}\sim\hspace{2pt}_K \psi$ iff $\hspace{1pt}\vdash\hspace{-6pt}\sim\hspace{2pt}_{Add(K,\varphi)} \psi$, where $Add(K, \varphi) = (\mathcal{F} \cup \{\varphi\}, \Delta)$ (see [1]).

[13]As shown in [23] and [1], the binary and prioritized version of $\hspace{1pt}\vdash\hspace{-6pt}\sim\hspace{2pt}_1$ is a rational inference relation which is furthermore well adapted to the handling of default rules.

**Definition 4** $K \hspace{2pt}\hspace{2pt}\mid\!\sim_2 \hspace{2pt} \psi$ iff $\exists \alpha > 0$ such that $Bel_K(\psi) \rightarrow_{\varepsilon \rightarrow 0} \alpha$

**Proposition 6** $K \mid\!\sim_2 \psi$ iff there is a maxcard scenario $S$ such that $S \models \psi$.

Again, the proof comes from the fact that the consistent environments with a non infinitely small belief are the maxcard ones. This CR is the *weak* (existential) counterpart of $\mid\!\sim_1$.

**Definition 5** $K \mid\!\sim_3 \psi$ iff $\exists \alpha > 0$ s.t. $Bel_K(\psi) \rightarrow_{\varepsilon \rightarrow 0} \alpha$ and $Bel_K(\neg\psi) \rightarrow_{\varepsilon \rightarrow 0} 0$.

**Proposition 7** $K \mid\!\sim_3 \psi$ iff $K \mid\!\sim_2 \psi$ and $K \not\mid\!\sim_2 \neg\psi$.

The proof comes straightforwardly from Propositions 5 and 6. This kind of CR, called *argumentative* in [2], is intermediate between weak and strong CRs.

**Definition 6** $K \mid\!\sim_4 \psi$ iff $Bel_K(\psi) > 0$

**Proposition 8** $K \mid\!\sim_4 \psi$ iff there is a scenario $S$ of $K$ such that $S \models \psi$, or equivalently, iff there is an irredundant scenario $S$ of $K$ such that $S \models \psi$.

The proof comes from the fact that all consistent environments have a non-zero probability. This well-known weak CR corresponds to the provability in at least one extension of a normal default theory without prerequisites[14].

**Definition 7** $K \hspace{2pt}\mid\!\sim_5 \hspace{2pt} \psi$ iff $Bel_K(\psi) > 0$ and $Bel_K(\neg\psi) = 0$

**Proposition 9** $K \mid\!\sim_5 \psi$ iff there is a scenario of $K$ entailing $\psi$ and if there is no scenario of $K$ entailing $\neg\psi$.

This result is a corollary of Proposition 8. This CR is another argumentative CR.

**Definition 8** $K \mid\!\sim_6 \psi$ iff $Bel_K(\psi) > \frac{1}{2}$

**Proposition 10** $K \mid\!\sim_6 \psi$ iff $\psi$ is provable in the majority (more than one half) of the maxcard scenarios of $K$.

The proof comes directly from the fact that all maxcard consistent environments have all the same probability (namely $\frac{1}{p}$) and that the non-maxcard ones have an infinitely small probability. This kind of CR has been called *majority* CR in [30].

**Definition 9** $K \mid\!\sim_7 \psi$ iff $Bel_K(\psi) > Bel_K(\neg\psi)$

---
[14]The corresponding strong CR (provability in all extensions), which is more interesting and which has received many improvements in the literature, seems to have no nice characterization in our framework.

A sufficient condition for $K \mid\!\sim_6 \psi$ to hold is when the number of maxcard scenarios entailing $\psi$ is greater than the number of maxcard scenarios entailing $\neg\psi$. However this condition is not necessary; the exact characterization is more complex:

**Proposition 11** Let $u(k, \psi)$ be the number of scenarios of $K$ of cardinality $k$ entailing $\psi$. Let $diff(\psi, \neg\psi) = Max\{k, u(k, \psi) \neq u(k, \neg\psi)\}$. Then $K \mid\!\sim_7 \psi$ iff $u(diff(\psi, \neg\psi), \psi) > u(diff(\psi, \neg\psi), \neg\psi)$.

Here is a sketch of the proof: there are 3 situations where $Bel_K(\psi) > Bel_K(\neg\psi)$:
- $Bel_K(\psi) = \frac{u_\psi}{p}$ and $Bel_K(\neg\psi) = \frac{u_{\neg\psi}}{p}$ where $u_\psi > u_{\neg\psi} > 0$; in this case, let $k^*$ be the cardinality of the maxcard consistent environments, then $u(k^*, \psi) = u_\psi$, $u(k^*, \neg\psi) = u_{\neg\psi}$ and $diff(\psi, \neg\psi) = k^*$;
- $Bel_K(\psi) = O(\varepsilon^{k_1})$ and $Bel_K(\neg\psi) = O(\varepsilon^{k_2})$ with $k_2 > k_1 \geq 0$, or $Bel_K(\neg\psi) = 0$; in this case, $u(k^* - k_1, \psi) > 0$, $u(k^* - k_1, \neg\psi) = 0$ and $diff(\psi, \neg\psi) = k^* - k_1$;
- $Bel_K(\psi) = O(\varepsilon^{k_1})$ and $Bel_K(\neg\psi) = O(\varepsilon^{k_1})$ as well; in this case, we have to develop further the expression of $Pr(E|\neg N)$ in Proposition 1, which would show that if among the consistent environments of cardinality $k_\psi(= k_{\neg\psi})$, there are more entailing $\psi$ than entailing $\neg\psi$, but if there are exactly as many, then it depends on the number of consistent environments of cardinality $k_\psi - 1$, and so on.

It is clear from this proof that this CR has a lexicographic spirit: indeed, Proposition 11 could have stated equivalently by:

**Proposition 12** Let $L(\psi) = (u(k, \psi), k = n...1)$ and $L(\neg\psi) = (u(k, \neg\psi), k = n...1)$ then $K \mid\!\sim_7 \psi$ iff $L(\psi) >_{lex} L(\neg\psi)$, where $>_{lex}$ is the lexicographic ordering.

**Definition 10** $K \hspace{2pt}\mid\!\sim_8 \hspace{2pt} \psi$ iff $Bel_K(\psi) > 0$ and $\frac{Bel_K(\neg\psi)}{Bel_K(\psi)} \rightarrow_{\varepsilon \rightarrow 0} 0$.

**Proposition 13** $K \mid\!\sim_8 \psi$ iff $k_\psi > k_{\neg\psi}$, where $k_\psi$ is been defined like in Proposition 4, with the convention $k_\psi = -\infty$ iff $\psi$ appears in the context of no consistent environment.

The proof comes easily from Proposition 4.

**Definition 11** $K \mid\!\sim_9 \psi$ iff $Bel_K(\psi) = 1$.

**Proposition 14** $K \mid\!\sim_9 \psi$ iff $\mathcal{F} \models \psi$.

This is a clone of Proposition 3 and has thus already been proved. This CR is very strong and it is even monotonic since it accepts only the consequences of hard facts.

**Proposition 15** Let $\prec$ be the relation between CR's (as in [30]) defined by $\mid\!\sim_i \prec \mid\!\sim_j$ iff $\forall K, \psi$, $K \mid\!\sim_i \psi \Rightarrow K \mid\!\sim_j \psi$. This relation between our $\mid\!\sim_i$'s is depicted by the graph on Figure 1.

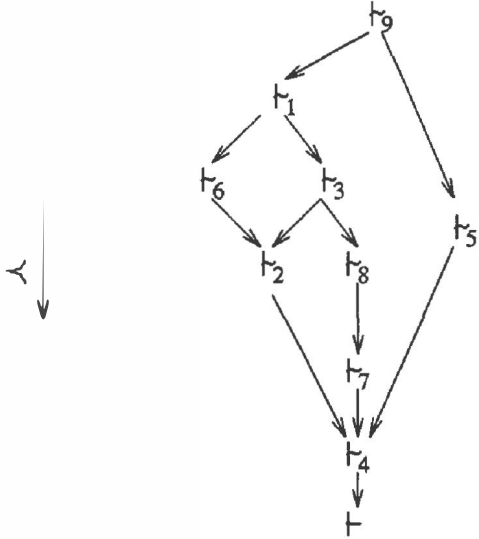

Figure 1: the $\prec$ relation between $\mathrel|\!\sim_i$'s

| | $\mathrel|\!\sim_1$ | $\mathrel|\!\sim_6$ | $\mathrel|\!\sim_3$ | $\mathrel|\!\sim_2$ | $\mathrel|\!\sim_8$ | $\mathrel|\!\sim_7$ | $\mathrel|\!\sim_5$ | $\mathrel|\!\sim_4$ |
|---|---|---|---|---|---|---|---|---|
| $a$ | × | × | × | × | × | × | × | × |
| $b \vee c$ | × | × | × | × | × | × | – | × |
| $c$ | – | × | × | × | × | × | – | × |
| $b$ | – | × | – | × | – | × | – | × |
| $f$ | – | – | × | × | × | × | × | × |
| $\neg d$ | – | – | – | × | – | – | – | × |
| $g$ | – | – | – | – | × | × | – | × |
| $\xi$ | – | – | – | – | × | × | × | × |
| $\neg g$ | – | – | – | – | – | – | – | × |
| $\neg f$ | – | – | – | – | – | – | – | – |

The proof would be long and tedious but does not present any particular difficulty. Note that the example at the end of the Section gives counterexamples corresponding to almost all couples of CRs such that $\mathrel|\!\sim_i \not\prec \mathrel|\!\sim_j$. Pinkas and Loui [30] define a *safe* CR as a CR $\mathrel|\!\sim$ such that $\forall K, \forall \psi, K \mathrel|\!\not\sim \psi$ or $K \mathrel|\!\not\sim \neg\psi$. It can be checked easily that $\mathrel|\!\sim_2$ and $\mathrel|\!\sim_4$ are generally unsafe while the other ones are safe.

**Proposition 16** *When $K$ is consistent, all $\mathrel|\!\sim_i$'s except $\mathrel|\!\sim_9$ collapse to classical entailment, i.e. $K \mathrel|\!\sim_i \psi$ iff $\mathcal{F} \cup \Delta \models \psi$.*

*Proof:* when $K$ is consistent, there is only one maxcard scenario: $K$ itself. Therefore $K \mathrel|\!\sim_1 \psi$ iff $K \models \psi$, and a fortiori, all $\mathrel|\!\sim_i$ below in the graph collapse to $\models$. It remains to show it for $\mathrel|\!\sim_5$, which is straightforward.

This list of CR's is of course not exhaustive and one could think of giving other definitions, possibly parametrized by a given $\alpha > 0$. The interest of such a list of CR's is to enable the user to use the most adapted CR to her specific problem, knowing that whichever CR she chooses, it will have an interpretation in terms of the belief function induced by the KB and assumptions (I) and (S). While very cautious CRs such as $\mathrel|\!\sim_1$ and very adventurous ones such as $\mathrel|\!\sim_4$ or $\mathrel|\!\sim_5$ are often considered as too extreme in practice, the more quantitative CRs $\mathrel|\!\sim_6$, $\mathrel|\!\sim_7$ (and $\mathrel|\!\sim_8$ which is maybe a bit less quantitative) seem to be good compromises inbetween, and furthermore their DS interpretation is appealing.

*Example (continued):*

Let $\xi = (b \vee c \vee \neg e) \wedge g$. In the following table, the sign × (resp. the sign –) means that the formula is (resp. is not) entailed w.r.t. $\mathrel|\!\sim_i$. There is no column for $\mathrel|\!\sim_9$ (obviously, only $a$ is a $\mathrel|\!\sim_9$-consequence of $K$).

## 5 Extension to the prioritized case

Many syntax-based approaches to default reasoning assume that the knowledge base is partitioned into priority levels, namely $K = (K_1, ..., K_n)$ (1 being by convention the most prioritary level); these levels are qualitative and generally it is more acceptable to violate any number of formulas of a lower priority level then violate one formula of a higher priority level. A generalization of the maximum cardinality principle to the prioritized case is defined both in [1] and in [23]: a sub-base $A$ of a prioritized knowledge base $(K_1, ..., K_n)$ is *lexicographically* strictly preferred to a sub-base $B$ iff there exists a $i \in 1..n$ such that $\forall j > i$, $|A \cap S_j| = |B \cap S_j|$ and $|A \cap S_i| > |B \cap S_i|$; the same selection criterium has been used in diagnosis by DeKleer [12]. Now, it is possible to characterize lexicographically preferred subtheories in terms of probabilities of fault; following De Kleer [12], for any piece of information in $K_i$ we assign an initial probability of fault of $\varepsilon_i$ to its source, with the constraint that $\forall i$ and $\forall j > i$, $\varepsilon_j \ll \varepsilon_i$ (more precisely, that $\varepsilon_j < \varepsilon_i^{fmax}$, where $fmax$ is an upper bound of the maximum number of formulas of a priority level - we may take for instance $fmax = |K|$). Then it can be shown that the only consistent environments of $K$ having an a posteriori probability which does not tend to 0 when $\varepsilon \to 0$ correspond exactly to the lexicographically preferred subbases (which generalizes Proposition 1), and that $\psi$ is lexicographically deduced from $K$ iff $Bel_K(\psi) \to_{\varepsilon \to 0} 1$ (which generalizes Proposition 5).

## 6 Related work and conclusion

We have strengthened the already known connections between syntax-based default reasoning, model-based diagnosis and ATMS, and belief functions, by building on deKleer's infinitesimal probabilities of fault. We have followed the following steps:

(1) syntax-based nonmonotonic entailment is viewed as diagnosis, by considering each piece of information as the description of how a component works, and the source which provided us with it as the component;

(2) we assume that all sources have very small (and equal) initial probabilities of fault, and that they are

independent;

(3) we compute the probabilities of each candidate, and then a belief function on the language which can be interpreted as a probability of provability;

(4) we use this belief function to define syntax-based nonmonotonic consequence relations;

(5) lastly, we position these definitions in the literature of syntax-based approaches to nonmonotonic entailment.

Our work integrates various subfields of AI and thus there are many rel;ated works, more so bacause the links between these subfliels had already received a lot of attention in the literature. Many authors assign prior probabilities to assumptions or components and compute then posterior probabilities of candidates, and some of them compute a belief function ([34] [27] [22], [35]) but generally the initial probabilities are assumed to be given by the user. De Kleer [12] uses the same basic assumptions as us (2) but he computes then posterior probabilities of candidates conditioned by a measurement (in order to find the best measurement to do next), which diverges from our step (4).

Furthermore, it is worth noticing that the "nontrivial" (i.e. other than 0 and 1) belief weights we obtain when $\varepsilon \rightarrow 0$ are obtained from a completely syntactic knowledge base without explicit numbers. A related work which shares this feature is the Dempster-Shafer handling of default rules of Benferhat and Smets [3]: they start from a set of ranked default rules, where the ranking comes either from the user or from the ranking procedure of Goldszmidt and Pearl's System Z [28]; they associate then to each default of rank $i$ the mass $1-\varepsilon_i$ (with $\varepsilon_i \ll 1$ and $\varepsilon_{i+1} \ll \varepsilon_i$) and compute then a belief function assuming independence between the defaults. Note that $\varepsilon_i^k$ and $\varepsilon_j^{k'}$, as well as arbitrary products of $\varepsilon_i$'s, are not comparable. The computed belief function is used to define an inference relation (in the same way as $\hspace{0.1em}\mid\!\sim_1$) which solves the property inheritance blocking problem. The common point to their work and ours is the generation of a belief function from a knowledge base (in their approach, a ranked set of default rules); but the objective pursued is different: while they search for a consequence relation solving the blocking inheritance problem, we want to characterize consequence relations in terms of probabilities of fault of the sources. Other authors have used infinitesimal probabilities in nonmonotonic reasoning, following Adams' $\varepsilon$-semantics (especially Pearl [27], Goldszmidt et al. [18]); in these approaches the default rule $\alpha \rightarrow \beta$ is translated by $Pr(\beta|\alpha) \geq 1 - \varepsilon$ with $\varepsilon \ll 1$. The main difference with our use of infinitesimal probabilities relies in their interpretation (in the latter approaches they are conditional probabilities qualifying default rules while in ours they qualify the relevance of a piece of information).

Obviously, steps (3) and (4) can be done without assuming infinitely small prior probabilities. Thus, the definitions given in Section 4 still make sense in the case we start with non-infinitesimal user-given probabilities of failure; but the results do not hold any longer and the characterization of these inference relations is thus much less interesting.

We would like to emphasize that our contribution does not really propose a new formalism nor a new way to perform nonmonotonic reasoning, but rather puts together the (already known) links between syntax-based default reasoning on one side, and ATMS, diagnosis and belief functions on the other side, and assumes further independence of the pieces of information and infinitely small probabilities of failure. Now, although the theoretical complexity of syntax-based entailment relations has received recently some attention [25] [7], up to now, the more practical algorithmic and implementation issues have been less studied in the literature of syntax-based default reasoning than in the literature of ATMS and model-based diagnosis. Therefore, our conclusion (and our hope) is that that syntax-based default reasoning should benefit from existing works on the aforementioned fields, such as the characterization of tractable subclasses (e.g. [5]), experimental results etc.

## Acknowledgements

This work has been supported by the ESPRIT-BRA Research Project "DRUMS-2" (Defeasible Reasoning and Uncertainty Management Systems). Thanks to Salem Benferhat, Didier Dubois and Henri Prade for helpful discussions and comments on earlier versions, and to the anonymous reviewers for helpful criticisms.